%% file: FinalTex.tex
\documentclass[journal]{IEEEtran}
\IEEEoverridecommandlockouts
\input{definition}

\def\BibTeX{{\rm B\kern-.05em{\sc i\kern-.025em b}\kern-.08em
    T\kern-.1667em\lower.7ex\hbox{E}\kern-.125emX}}
\usepackage{makecell}

\begin{document}

\title{Proprioceptive Invariant State Estimation for Humanoid Robots on Non-Inertial Ground}

\author{Falak Mandali$^{1,*}$, Zijian He$^{1,*}$, Yan Gu$^{1,\dagger}$
\thanks{$^{1}$Purdue University, West Lafayette, IN 47907, USA.
	E-mails: {\{\tt \footnotesize fmandali,he348,yangu\}}@purdue.edu.}
\thanks{$^*$Equal contributions. $^\dagger$Corresponding author.}
\thanks{This research has been supported in part by NSF CMMI-2423239 and ONR N00014-24-1-2028.}
}

\maketitle

\begin{abstract}
This paper presents an invariant extended Kalman filtering (InEKF) approach for real-time state estimation of humanoid robots operating on non-inertial ground using only onboard proprioceptive sensing. The proposed approach estimates the robot’s base position and velocity relative to the moving ground frame without requiring direct measurements of ground motion or externally mounted sensors. By exploiting kinematic constraints at the stance foot through foot-mounted IMUs, the filter accounts for ground-induced nonlinearities in the process and measurement models while remaining fully proprioceptive. The estimator is formulated to admit a right-invariant measurement model, enabling favorable error dynamics under large initial uncertainties. Observability analysis establishes conditions under which the robot’s relative base position and velocity are observable with respect to the non-inertial ground frame. Experiments with the Digit humanoid robot standing and squatting atop a swaying and pitching ground showcase a 96\% speedup in convergence rate and an 80\% reduction in position estimate errors over existing InEKFs. Walking experiments on a uni-axially rotating ground achieve an average estimation error of less than 9 cm for an initial error of up to 1 m.

\end{abstract}


\vspace{-0.15in}
\section{Introduction} \label{sec:intro}
\vspace{-0.05in}

Accurate, real-time state estimation is a prerequisite for model-based planning and control of legged robots. Most existing estimators fuse proprioceptive and exteroceptive sensing using extended Kalman filters (EKFs) \cite{bloesch_state_2012} or optimization-based frameworks \cite{camurri_pronto_2020, cheng2025simultaneous}, and rely on the assumption that the ground is static, implying zero stance-foot velocity in the inertial frame \cite{bloesch_state_2012, agrawal_proprioceptive_2022}.
Under this assumption, invariant EKFs (InEKFs) \cite{barrau_invariant_2015}, which ensure trajectory-independent linearization under group-affine dynamics and invariant observation conditions, have significantly improved estimation accuracy and convergence rate beyond standard EKF methods \cite{hartley_contact-aided_2019,teng_legged_2021,gao_invariant_2022,he_invariant_2025,lin2021legged}.
In parallel, recent proprioceptive estimation methods have explored the use of multiple onboard inertial measurement units (IMUs) to mitigate drift or improve vertical-position estimation \cite{wu2025doglegs, yang2025multi, xavier_multi-imu_2023}, all of which fundamentally rely on the static-ground assumption.

However, the static-ground assumption 
limits applicability in non-inertial environments, where the supporting surface undergoes translational and rotational motion relative to the inertial frame, such as on moving trains, ships, or aircraft \cite{gao2025time,iqbal2020provably, stewart2025adaptive,iqbal2023analytical}. In these settings, ground-induced accelerations and rotations fundamentally alter both the system dynamics and the structure of the estimation problem, where ignoring ground motion introduces significant modeling errors that degrade estimation accuracy, particularly under large ground motion.

\begin{figure}
    \centering
    \includegraphics[width=1\linewidth]{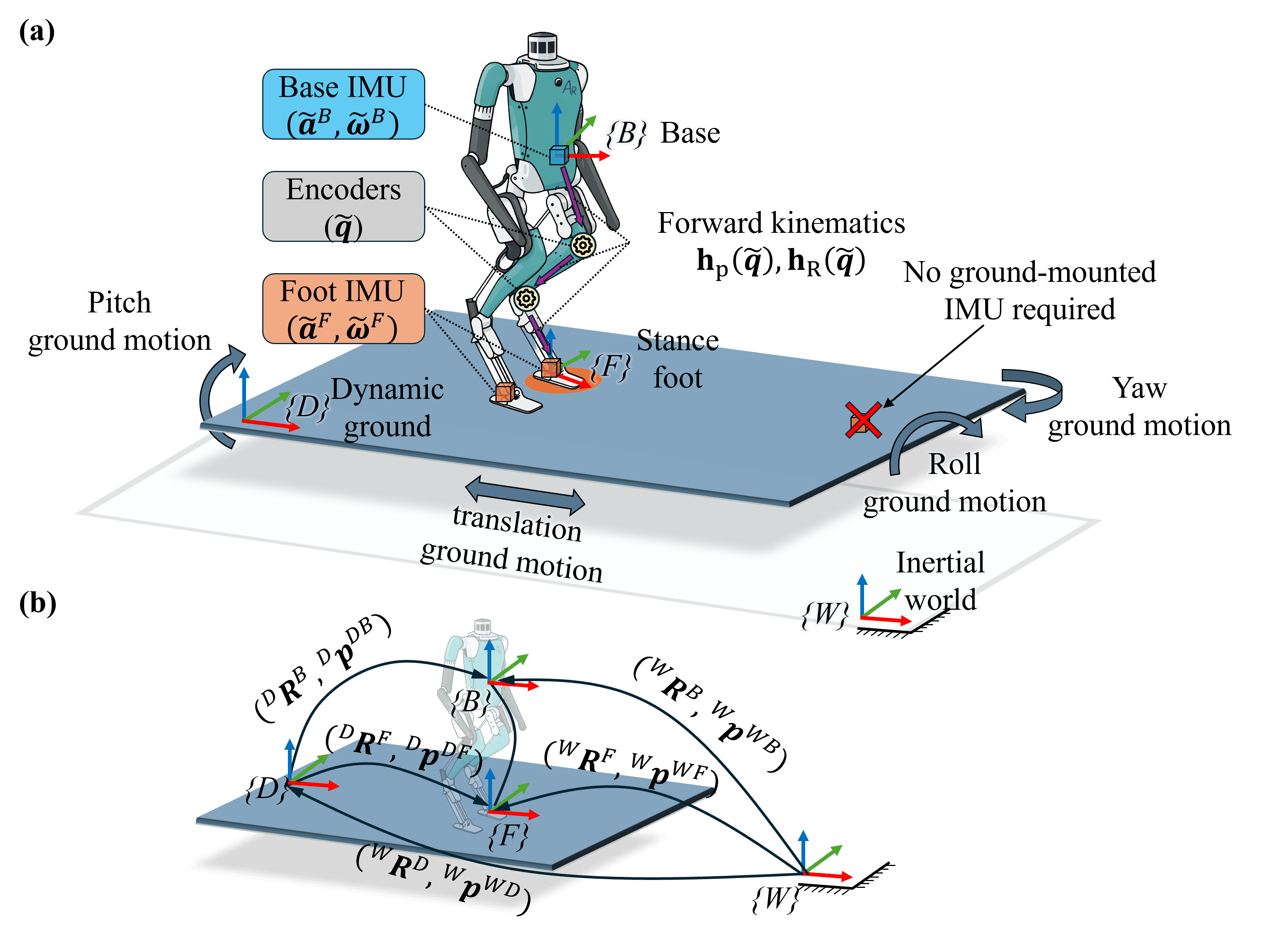}
    \vspace{-0.3in}
    \caption{(a) Humanoid robot walking on non-inertial ground with onboard proprioceptive sensors (torso and foot IMUs and joint encoders); no ground-mounted sensors are required. (b) Reference frames and state variables used for state estimation.}
    \label{fig:problem setting}
    \vspace{-0.25in}
\end{figure}

To address state estimation on non-inertial ground, the InEKF introduced in \cite{gao_invariant_2022} explicitly models ground motion to estimate a robot’s absolute state in the inertial world frame, but assumes direct and accurate knowledge of the moving ground’s absolute pose and velocity, which is an assumption rarely satisfied in practice. More recently, \cite{he_legged_2024, he_invariant_2025} formulate state estimation in a non-inertial frame rigidly attached to the moving ground, enabling full observability of the robot’s relative pose and velocity under general ground motions. Yet, this approach relies on an external IMU physically mounted on the ground to measure its motion, which may be impractical in real-world deployments due to instrumentation constraints, communication delays, and potential signal loss, particularly in unstructured environments. 
Recent learning-based~\cite{lee2026attention, yu2024state} and hybrid model-learning~\cite{11127971, lin2021legged} estimators have shown promising results under contact uncertainty, slip, and complex terrain. Yet, they rely on substantial training data and estimate states in the inertial frame, so their generalization to non-inertial environments remains unclear.

To enable accurate, real-time state estimation of humanoid robots in non-inertial environments without explicit ground motion sensing, we propose an InEKF that estimates the robot’s motion state relative to a moving ground frame using only onboard proprioceptive sensors. These state estimates are directly relevant for planning and control tasks defined with respect to the moving environment.
The key contributions of this study are summarized as follows:
\begin{enumerate}
    \item [(a)] 
    An InEKF formulation for estimating a humanoid robot’s base pose and velocity relative to a non-inertial ground frame using only joint encoders and torso- and foot-mounted IMUs. By exploiting kinematic foot-ground contact constraints through foot-IMU measurements, the proposed InEKF enables fully proprioceptive estimation.
    \item [(b)] 
    Process and measurement models that explicitly account for nonlinear effects induced by ground motion, with the measurement model satisfying the right-invariant observation condition. These properties promote robust and accurate estimation under large initial errors.
    \item [(c)] An observability analysis revealing that the base position and velocity relative to the moving ground are observable under general ground motions. The analysis further shows that simple augmentation of the sensing configuration enables recovery of base orientation observability.
    \item [(d)] Hardware experiments on the Digit humanoid robot under various ground motions, demonstrating improved estimation accuracy and faster convergence of the proposed InEKF compared to baseline estimators, and validating the theoretical observability results.
    
\end{enumerate}

\vspace{-0.1 in}
\section{Preliminaries} \label{sec:prelim}

We consider a matrix Lie group $\mathcal{G} \subset \mathrm{GL}(n, \mathbb{R})$, where $\mathrm{GL}(n, \mathbb{R})$ denotes the set of invertible $n \times n$ real matrices. Its associated Lie algebra, denoted by $\mathfrak{g}$ with dimension $\mathrm{dim}(\mathfrak{g})$, acts as the tangent space of $\mathcal{G}$ at the identity element $\mathbf{I} \in \mathcal{G}$. We define the wedge operator $(\cdot)^{\wedge}$: $\mathbb{R}^{\mathrm{dim}(\mathfrak{g})} \rightarrow \mathfrak{g}$ as the isomorphism that maps a vector $\boldsymbol{\xi} \in \mathbb{R}^{\mathrm{dim}(\mathfrak{g})}$ to the corresponding element in the Lie algebra. The exponential map {$\exp: \mathbb{R}^{\mathrm{dim (\mathfrak{g})}} \rightarrow \mathcal{G}$} is defined as
{\small $ \label{matrix exp}
    \exp\left( \boldsymbol{\xi}\right) = \exp_{m}\left( \boldsymbol{\xi}^{\wedge}\right)$},
where {$\exp_{m}(\cdot)$} is the matrix exponential.
The Adjoint map $\mathbf{Ad}_{\mathbf{X}}:\mathfrak{g} \rightarrow \mathfrak{g}$ at $\mathbf{X}\in \mathcal{G}$ is defined as $\left(\mathbf{Ad}_{\mathbf{X}} \boldsymbol{\xi}\right)^{\wedge} = \mathbf{X}\boldsymbol{\xi}^{\wedge} \mathbf{X}^{-1}$.

Throughout this paper, we denote $\mathbf{I}_{n}$ and $\mathbf{0}_{m,n}$ as the $n \times n$ identity matrix and the $m \times n$ zero matrix, respectively. The skew-symmetric matrix associated with a vector $(\cdot) \in \mathbb{R}^{3}$ is denoted by $[\cdot]_{\times}$. We distinguish between the true values of a variable $(\cdot)$, its measured value $\tilde{(\cdot)}$, and its estimated value $\bar{(\cdot)}$. $(\cdot)_t$ denotes the value of a variable $(\cdot)$ at time $t$. 

We define four primary reference frames: the robot's base (e.g., torso) link $\{B\}$, the stance foot $\{F\}$, the dynamic ground $\{D\}$, and the inertial world $\{W\}$, as illustrated in Fig.~\ref{fig:problem setting}(a).

We adopt the convention that the left superscript denotes the frame in which a quantity is expressed, while the right superscript specifies the frame relationship. A single-letter right superscript indicates the frame whose configuration is being expressed (e.g., ${}^{D}\mathbf{R}^{B}$ is the orientation of ${B}$ relative to ${D}$), whereas in a two-letter right superscript, the first denotes the reference origin and the second denotes the target frame (e.g., ${}^{W}\mathbf{p}^{DB}$ is the position of the origin of ${B}$ relative to ${D}$, expressed in the world frame ${W}$).
The reference frames and variables are illustrated in Fig.~\ref{fig:problem setting}(b). Table I in the supplementary material summarizes the main notations.

\vspace{-0.13 in}
\section{Problem Formulation} \label{sec:problem}
\vspace{-0.08 in}


\subsection{State Representation}
\vspace{-0.05 in}
For reliable navigation in non-inertial environments, planners and controllers require the robot’s motion state relative to the moving environment rather than an inertial world frame, since task objectives such as obstacle avoidance and goal reaching are naturally defined relative to the moving support. Accordingly, the proposed filter estimates the robot's velocity and pose with respect to the dynamic ground frame  $\{D\}$.

Following the reference frame conventions in Sec.~\ref{sec:prelim}, the robot base pose and velocity in the inertial frame $\{W\}$ are denoted by ${}^{W}\mathbf{R}^{B}_{t}$,${}^{W}\mathbf{p}^{WB}_{t}$, and ${}^{W}\mathbf{v}^{WB}_{t}$, with the stance-foot position given by ${}^{W}\mathbf{p}^{WF}_{t}$.
The ground motion relative to $\{W\}$ is parameterized by ${}^{W}\mathbf{R}^{D}_{t}$, ${}^{W}\mathbf{v}^{WD}_{t}$, and ${}^{W}\mathbf{p}^{WD}_{t}$.
The robot’s motion relative to the moving ground is described by ${}^{D}\mathbf{R}^{B}_{t}$ and ${}^{D}\mathbf{p}^{DB}_{t}$, together with the stance-foot pose ${}^{D}\mathbf{R}^{F}_{t}$ and ${}^{D}\mathbf{p}^{DF}_{t}$, as illustrated in Fig.~\ref{fig:problem setting}(b).

\subsubsection{State variables}

To estimate the robot’s motion relative to the dynamic ground frame $\{D\}$, we define the relative base orientation, position, and velocity as:
\begin{equation} \label{eq:relative pose}
\small 
\begin{aligned}
\vspace{-0.05 in}
        \mathbf{R}_{t}^{B} &:= {}^{D}\mathbf{R}^{B}_{t} = ({}^{W}\mathbf{R}^{D}_{t})^{\transpose} ({}^{W}\mathbf{R}^{B}_{t}),
        \\
        \mathbf{p}_{t} &:= {}^{D}\mathbf{p}^{DB}_{t} = ({}^{W}\mathbf{R}^{D}_{t})^{\transpose} ({}^{W}\mathbf{p}^{WB}_{t} - {}^{W}\mathbf{p}^{WD}_{t}),
        \\
        \mathbf{v}_{t} &:= ({}^{W}\mathbf{R}^{D}_{t})^{\transpose} ({}^{W}\mathbf{v}^{WB}_{t} - {}^{W}\mathbf{v}^{WD}_{t}),
\end{aligned}
\end{equation}
where ${}^{W}\mathbf{v}^{WB}_{t} = \frac{d}{dt}({}^{W}\mathbf{p}^{WB}_{t})$ and {${}^{W}\mathbf{v}^{WD}_{t} = \frac{d}{dt}({}^{W}\mathbf{p}^{WD}_{t})$} are time derivatives taken in the inertial world frame $\{ W\}$.
Note that $\mathbf{v}_t$ is not equal to $\frac{d}{dt}\mathbf{p}_{t}$ when $\{D\}$ is rotating relative to $\{ W\}$, as the latter additionally includes terms induced by the angular motion of $\{D\}$.

To exploit stance-contact kinematic constraints and eliminate direct ground-motion sensing, we further introduce the stance-foot orientation and position relative to $\{D\}$ as:
\begin{equation} \label{eq:relative pose foot}
\small 
\begin{aligned}
\vspace{-0.05 in}
        \mathbf{R}_{t}^{F} &:= {}^{D}\mathbf{R}^{F}_{t} = ({}^{W}\mathbf{R}^{D}_{t})^{\transpose} ({}^{W}\mathbf{R}^{F}_{t}),
        \\
        \mathbf{d}_{t} &:= {}^{D}\mathbf{p}^{DF}_{t} = ({}^{W}\mathbf{R}^{D}_{t})^{\transpose} ({}^{W}\mathbf{p}^{WF}_{t} - {}^{W}\mathbf{p}^{WD}_{t}).
\end{aligned}
\end{equation}

\subsubsection{State representation on a Lie group}
The state variables in \eqref{eq:relative pose} and \eqref{eq:relative pose foot} are embedded in a matrix Lie group $\mathcal{G}$ as: 
\begin{equation}
\small
\label{eq:right-error}
\mathbf{X}_t :=
    \begin{bmatrix}
        {\mathbf{R}}^F_t & \mathbf{v}_t &\mathbf{p}_t &\mathbf{d}_{t} &\mathbf{0}_{3,3}
        \\
        \mathbf{0}_{1,3} & 1 & 0 & 0 & \mathbf{0}_{1,3}
        \\
        \mathbf{0}_{1,3} & 0 & 1 & 0 & \mathbf{0}_{1,3}
        \\
        \mathbf{0}_{1,3} & 0 & 0 & 1 & \mathbf{0}_{1,3}
        \\
        \mathbf{0}_{3,3} &\mathbf{0}_{3,1} &\mathbf{0}_{3,1} &\mathbf{0}_{3,1} &\mathbf{R}_t^B
    \end{bmatrix} \in \mathcal{G},
\end{equation}
where $\mathcal{G}$ is composed of a double spatial isometry $SE_3(3)$ and the special orthogonal group $SO(3)$. 

\subsubsection{Right-invariant and logarithmic errors}
By the general InEKF methodology~\cite{barrau_invariant_2015}, 
both the propagation and update steps are formulated in terms of the right-invariant error $\boldsymbol{\eta}_t \in \mathcal{G}$ and its associated logarithmic error {\small $\boldsymbol{\xi}_{t} \in \mathbb{R}^{\text{dim}\mathfrak{g}}$}.
The right-invariant error, defined as $\boldsymbol{\eta}_t := \bar{\mathbf{X}}_t \mathbf{X}_t^{-1}$, captures the discrepancy between the true state $\mathbf{X}_t$ and its estimate $\bar{\mathbf{X}}_t$. The logarithmic error satisfies $\boldsymbol{\eta}_{t} = \exp(\boldsymbol{\xi}_{t})$, with
{\small $\begin{bmatrix}
    (\boldsymbol{\xi}^{R^F}_{t})^{\transpose}
    ,
    (\boldsymbol{\xi}^{v}_{t})^{\transpose}
    ,
    (\boldsymbol{\xi}^{p}_{t})^{\transpose}
    ,
    (\boldsymbol{\xi}^{d}_{t})^\transpose
    ,
    (\boldsymbol{\xi}^{R^B}_{t})^{\transpose}
\end{bmatrix}^\transpose:=\boldsymbol{\xi}_{t}$}. 


\vspace{-0.15in}
\subsection{Sensors Considered}
\vspace{-0.05in}
\label{subsec:sensor}

To ensure practical deployability, the proposed filter relies exclusively on onboard proprioceptive sensing and does not require external instrumentation (e.g., ground-mounted IMUs) or exteroceptive sensors such as GPS, LiDAR, or cameras. Specifically, the filter utilizes onboard IMUs and joint encoders commonly available on legged robots.

\subsubsection{Proprioceptive sensor selection} 

Encoders measure all joint angles, and each IMU provides linear acceleration and angular velocity measurements expressed in the IMU frame. We consider three IMUs mounted on the robot: one on the base (torso) and one on each footpad, as illustrated in Fig.~\ref{fig:problem setting}(a).

\subsubsection{Reference frame assignment}

For simplicity and without loss of generality, we assume that (i) during double-support or standing phases, the stance-foot IMU data from either foot may be used, and (ii) the robot base frame $\{B\}$ and stance-foot frame $\{F\}$ coincide with the frames of the base IMU and stance-foot IMU, respectively.

\subsubsection{Sensor data expression}

Sensor measurements are modeled as true values corrupted by additive white Gaussian noise. Although real sensors may exhibit colored noise and slowly varying bias drift, these effects are not explicitly modeled here and are instead mitigated through offline calibration and covariance tuning. Accordingly, the joint angle measurements satisfy $\tilde{\mathbf{q}}_t = \mathbf{q}_t + \mathbf{w}^q_t$, where $\mathbf{w}^q_t$ denotes encoder noise. 

Let ${}^{i}\tilde{\boldsymbol{\omega}}^{Wi}_t$ and ${}^{i}\tilde{\mathbf{a}}^{Wi}_{t}$ denote the measured angular velocity and linear acceleration of frame $\{i\}$ with respect to the initial world $\{W\}$, expressed in frame $\{ i \}$, where $i\in \{B, F\}$. For brevity, we define $ {}^{i}\Tilde{\boldsymbol{{\omega}}}_{t}:={}^{i}\Tilde{\boldsymbol{{\omega}}}_{t}^{Wi}~\text{and}~{}^{i}\Tilde{\mathbf{{a}}}_{t}:={}^{i}\Tilde{\mathbf{{a}}}^{Wi}_{t}$, 
and model the measurements as ${}^{i}\Tilde{\boldsymbol{{\omega}}}_{t} = {}^{i}\boldsymbol{{\omega}}^{Wi}_{t} + {}^{i}\mathbf{w}_{t}^{\omega}~\text{and}~{}^{i}\Tilde{\mathbf{{a}}}_t = {}^{i}\mathbf{{a}}^{Wi}_{t} +  {}^{i}\mathbf{w}_{t}^{a},$ where ${}^i {\boldsymbol{\omega}}^{Wi}_t$ and ${}^i {\mathbf{a}}^{Wi}_t$ are the true angular velocity and linear acceleration, and ${}^i\mathbf{w}_{t}^{\omega}$ and ${}^i\mathbf{w}_{t}^{a}$ are white Gaussian noise.

\vspace{-0.15in}
\section{System Modeling} \label{sec:model}
\vspace{-0.05in}

This section derives the process and measurement models of the proposed filter. The process model governs the time evolution of the system state and forms the prediction step of the InEKF (Sec.~\ref{sec:inekf}-A), while the measurement model maps available proprioceptive sensor readings to the state for correction during the update step (Sec.~\ref{sec:inekf}-B).

As legged locomotion involves intermittent ground contact, the models account for both continuous swing phases and discrete contact events. Also, the translational and rotational motion of the non-inertial ground introduces nonlinearities, which are explicitly captured in the proposed modeling.

\vspace{-0.15 in}
\subsection{Process Model during a Continuous Phase}\label{subsec: process model}
\vspace{-0.05 in}

During continuous phases, the process model is constructed from IMU motion dynamics and stance-foot contact kinematics, which is simple yet accurate.

\subsubsection{Relative IMU motion dynamics} \label{sec: sys-process-rel}

We derive the relative IMU motion dynamics by expressing the robot’s motion in the dynamic ground frame $\{D\}$. Building on prior extensions of IMU dynamics from inertial frames to non-inertial frames \cite{he_invariant_2025}, the continuous-time evolution of the relative base orientation $\Rt[B]$, velocity $\mathbf{v}_t$, and position $\mathbf{p}_t$ is given by:
\begin{equation} \label{eq:process_model_he_2024}
\small
\begin{split}
    \tfrac{d}{dt}(\Rt[B] ) &= \Rt[B] [{}^{B}{\boldsymbol{{\omega}}}_{t} ]_\times - [{}^{D} {\boldsymbol{{\omega}}}_{t} ]_\times \Rt[B],
    \\
    \tfrac{d}{dt}(\mathbf{v}_t ) 
    &= -[{}^{D}{\boldsymbol{{\omega}}}_{t} ]_\times 
        \mathbf{v}_{t} + \mathbf{R}_{t}^{B} ({}^{B}{\boldsymbol{{\mathbf{a}}}}_{t} )  - {}^{D}{\mathbf{a}}_{t}  ,
    \\
    \tfrac{d}{dt}(\mathbf{p}_t) &= -[{}^{D}{\boldsymbol{{\omega}}}_{t}  ]_\times 
    \mathbf{p}_{t} + \mathbf{v}_{t}.
\end{split}
\end{equation}

Equation \eqref{eq:process_model_he_2024} depends on the dynamic ground acceleration {\small ${}^{D}{\mathbf{a}}_t$} and angular velocity {\small ${}^{D}{\boldsymbol{\omega}}_t$}, which are assumed to be directly measured by a ground-mounted IMU in prior works \cite{he_invariant_2025}. Since such measurements are generally unavailable without external instrumentation, we eliminate this dependency by exploiting stance-foot kinematic constraints.

As described in Sec.~\ref{subsec:sensor}, the robot feet are equipped with IMUs measuring {\small ${}^{F}\tilde{\mathbf{a}}_t$} and {\small ${}^{F}\tilde{\boldsymbol{\omega}}_t$}. Our objective is to express the ground-frame inertial quantities using only onboard proprioceptive sensing.

During stance contact, foot slippage and rolling are treated as noise (see Sec.~\ref{sec:Stance-foot contact dynamics}). In the absence of such effects, the relative stance-foot position and orientation satisfy {\small $\tfrac{d}{dt}(\mathbf{d}_t) = \mathbf{0}$} and {\small $\tfrac{d}{dt}(\mathbf{R}^F_t) = \mathbf{0}$}. Under these conditions, differentiating \eqref{eq:relative pose foot} yields expressions for the dynamic ground acceleration and angular velocity in terms of foot-IMU measurements:
\vspace{-0.05 in}
\begin{equation} \label{eq:DtoF_p}
\small
    {}^{D}\boldsymbol{{\mathbf{a}}}_t 
    = \Rt[F] ({}^{F}\boldsymbol{{\mathbf{a}}}^{WF}_{t}) -\Rt[F] \left([ {}^{F}\Dot{\boldsymbol{{\omega}}}^{WF}_t]_\times +  [ {}^{F}\boldsymbol{{\omega}}^{WF}_{t}]_\times^{2} \right) (\Rt[F])^{\transpose} \mathbf{d}_t,
\end{equation}
\begin{equation} \label{eq:DtoF_R}
\small
    [{}^{D}\boldsymbol{{\omega}}_{t}]_\times  = [(\Rt[F]){}^{F}\boldsymbol{{\omega}}^{WF}_{t}]_\times,
\vspace{-0.05 in}
\end{equation}
whose detailed derivations are in the supplementary file.

Introducing the forward-kinematics mapping {\small $\mathbf{h}_{R}(\mathbf{q}_t) = {}^{B}\mathbf{R}^{F}_t $} and substituting \eqref{eq:DtoF_p} and \eqref{eq:DtoF_R} in \eqref{eq:process_model_he_2024} yields a process model that no longer depends on direct ground-motion sensing:
\begin{equation}  \label{eq:proc_model}
\small
    \begin{aligned}
    \tfrac{d}{dt}\Rt[B] =& \Rt[B][\gyro[B]]_\times 
    \\
        &- \Rt[B][ (\mathbf{h}_{R}(\tilde{\mathbf{q}}) - \mathbf{J}_{h_R}(\tilde{\mathbf{q}}_t,  \mathbf{w}_{t}^{q}) ) (\gyro[F]) ]_\times,
    \\
        \tfrac{d}{dt}\mathbf{v}_t 
        =& -\Rt[F][\gyro[F]]_\times (\Rt[F])^{\transpose} 
        \mathbf{v}_t 
        \\
        &+ \Rt[F] (\mathbf{h}_{R}(\tilde{\mathbf{q}})^{\transpose} - \mathbf{J}_{h_R}(\tilde{\mathbf{q}}_t,  \mathbf{w}_{t}^{q})^{\transpose} ) (\acc[B]) 
        \\
        &- \Rt[F] (\acc[F]) 
        \\
        & + \Rt[F] \left( [ \gyro[F]]_\times^2 + [{}^{F}\dot{\tilde{\boldsymbol{{\omega}}}}_{t} - \gyroW[F]]_\times \right) (\Rt[F])^{\transpose}  \mathbf{d}_t,
    \\
    \tfrac{d}{dt}\mathbf{p}_t =& -\Rt[F][\gyro[F]]_\times (\Rt[F])^{\transpose} \mathbf{p}_t +  \mathbf{v}_t,
    \end{aligned}
\end{equation}
where 
the vector {\small $\mathbf{J}_{h_R}(\tilde{\mathbf{q}_t}, \mathbf{w}^q_t)$} is defined as
{\small $\mathbf{J}_{h_R}(\tilde{\mathbf{q}_t}, \mathbf{w}^q_t) :=\begin{bmatrix}
    \frac{\partial \mathbf{h}_{R,1}}{\partial \mathbf{q}_t} (\tilde{\mathbf{q}}_t) \mathbf{w}_{t}^{q}, & \frac{\partial \mathbf{h}_{R,2}}{\partial \mathbf{q}_t} (\tilde{\mathbf{q}}_t) \mathbf{w}_{t}^{q}, & \frac{\partial \mathbf{h}_{R,3}}{\partial \mathbf{q}_t} (\tilde{\mathbf{q}}_t) \mathbf{w}_{t}^{q}
\end{bmatrix}$},
which is obtained using the Jacobian of each column of 
{\small $\mathbf{h}_R =: \begin{bmatrix}
    \mathbf{h}_{R,1}, &\mathbf{h}_{R,2}, &\mathbf{h}_{R,3}
\end{bmatrix}$}. 
The physical interpretation of Eq. \eqref{eq:proc_model} is provided in the supplementary material.

\subsubsection{Stance-foot contact dynamics}
\label{sec:Stance-foot contact dynamics}

To model slippage, the foot velocity in $\{D\}$ is treated as white Gaussian noise. Then, the foot position {\small $\mathbf{d}_t$} and orientation {\small $\mathbf{R}^F_t$} evolve as:
\vspace{-0.05 in}
\begin{equation} \label{eq:foot dynamics}
\vspace{-0.05 in}
\small
    \tfrac{d}{dt} \mathbf{d}_{t} = \Rt[F] (-{}^{F}\mathbf{w}_{t}^{d} ), \quad
    \tfrac{d}{dt}  \mathbf{R}^F_t =  \Rt[F] [{}^{F}\mathbf{w}_{t}^{R^F} ]_\times,
\end{equation}
where ${}^{F}\mathbf{w}_{t}^{d}$ and ${}^{F}\mathbf{w}_{t}^{R^F} $ are white Gaussian noises.

\subsubsection{Overall process model}
Combining \eqref{eq:proc_model} and \eqref{eq:foot dynamics} gives: 
\vspace{-0.05 in}
\begin{equation}
\vspace{-0.05 in}
\small 
    \frac{d}{dt}(\mathbf{X}_t) = f_{u_t}(\mathbf{X}_t) + \mathbf{W}_t(\mathbf{X}_t, \mathbf{w}_t),
\end{equation}
where {\small$\mathbf{u}_t := [ ({}^{B}\tilde{\mathbf{a}}_t)^{\transpose}, ({}^{F}\tilde{\mathbf{a}}_t)^{\transpose}, ({}^{B}\tilde{\boldsymbol{\omega}}_t)^{\transpose}, ({}^{F}\tilde{\boldsymbol{\omega}}_t)^{\transpose} ]^{\transpose}$}
collects the IMU measurements and {\small $\mathbf{w}_t$} aggregates sensor noise and contact uncertainties as 
{\small $\mathbf{w}_t :=[({}^{B}\mathbf{w}_t^{a})^{\transpose} , 
({}^{F}\mathbf{w}_t^{a})^{\transpose} , ({}^{B}\mathbf{w}_t^{\omega})^{\transpose} , $ $({}^{F}\mathbf{w}_t^{\omega})^{\transpose} , (\mathbf{w}_t^q)^{\transpose} , ({}^{F}\mathbf{w}_t^{v})^{\transpose}, ({}^{F}\mathbf{w}_t^{\omega_c})^{\transpose} ]^{\transpose}$}.
The expressions of $f_{u_t}$ and $\mathbf{W}_t$ are given in the supplementary file.

\vspace{-0.15 in}
\subsection{Measurement Model} \label{subsec: measurement model}
\vspace{-0.05 in}

During continuous phases, we employ leg odometry as the measurement model. Based on the relative position between the base and stance-foot frames, the leg-odometry measurement is given by:
{\small $\mathbf{h}_p(\mathbf{q})=(\mathbf{R}^{B}_t)^{\transpose}(\mathbf{d}_t - \mathbf{p}_t)$}.
To express this measurement in the stance-foot frame used in the process model, we left-multiply by $\mathbf{h}_R^\transpose$, yielding:
\vspace{-0.05 in}
\begin{equation} \label{eq:FK}
\small
    \mathbf{h}_{R}^{\transpose} ({\mathbf{q}}_t) \mathbf{h}_p(\mathbf{{q}}_t) = (\mathbf{R}^{F}_t)^{\transpose}(\mathbf{d}_t - \mathbf{p}_t).
\vspace{-0.05 in}
\end{equation}

Due to encoder noise, the true measurement is approximated via first-order Taylor expansion as:
{\small $\mathbf{h}_{R}^{\transpose}(\mathbf{q}_t) \mathbf{h}_{p}({\mathbf{q}}_{t})
\approx 
\mathbf{h}_{R}^{\transpose}(\tilde{\mathbf{q}}_t)\mathbf{h}_{p}(\tilde{\mathbf{q}}_{t} ) - \mathbf{h}_{R}^{\transpose} \left(\mathbf{J}_{h_p}(\tilde{\mathbf{q}}_t) + [\mathbf{h}_{p}(\tilde{\mathbf{q}}_t)]_{\times} \mathbf{J}_{\theta}(\tilde{\mathbf{q}}_t) \right)  \mathbf{w}_{t}^{q} $},
where {\small $\mathbf{J}_{h_p}(\tilde{\mathbf{q}}_t) = \frac{\partial \mathbf{h}_{p}}{\partial\mathbf{q}} (\tilde{\mathbf{q}}_t)$} is the linear Jacobian and
{\small $\mathbf{J}_{\theta}(\tilde{\mathbf{q}}_t) \in \mathbb{R}^{3 \times n}$} is the angular component of the geometric Jacobian~\cite{lynch2017modern} mapping the joint velocities to the relative angular velocity of the foot.
Derivations are included in the supplementary file. 

Following invariant filtering theory, the measurement in \eqref{eq:FK} admits the right-invariant observation form \cite{barrau_invariant_2015}:
\vspace{-0.05 in}
\begin{equation}
\vspace{-0.05 in}
\label{eq:measurement}
\small
     \mathbf{Y}_t = \mathbf{X}_t^{-1} \mathbf{b} + \mathbf{n}_t, 
\end{equation}
where {\small $\mathbf{Y}_t = 
    \begin{bmatrix}
          (\mathbf{h}_R^\transpose (\tilde{\mathbf{q}}_t) \mathbf{h}_p(\tilde{\mathbf{q}}_t))^\transpose , &0, &1, &-1, &\mathbf{0}_{1,3}
    \end{bmatrix}^{\transpose}$}, 
{\small $\mathbf{n}_t = 
\begin{bmatrix}
    (\mathbf{h}_{R}^{\transpose}(\tilde{\mathbf{q}}_t)
\left(\mathbf{J}_{h_p}(\tilde{\mathbf{q}_t}) + [\mathbf{h}_{p}(\tilde{\mathbf{q}}_t)]_{\times} \mathbf{J}_{\theta}(\tilde{\mathbf{q}_t})\right) \mathbf{w}^{q}_{t} )^\transpose, & \mathbf{0}_{1,6}
\end{bmatrix}^\transpose$}, and 
{\small $\mathbf{b} =
\begin{bmatrix}
    \mathbf{0}_{1,3}, &0, &1, &-1, &\mathbf{0}_{1,3}
\end{bmatrix}^{\transpose}$}.
The derivations of {\small $\mathbf{Y}_t$}, {\small $\mathbf{b}$}, and {\small $\mathbf{n}_t$} are given in the supplementary file.

This invariant structure yields a log-error dynamics that is exactly linear and trajectory-independent in the deterministic case, enabling consistent covariance propagation, fast convergence, and high accuracy even under large initial errors~\cite{barrau_invariant_2015}.

\vspace{-0.15 in}
\subsection{Process Model at a Discrete Contact-Switch Event}
\vspace{-0.05 in}

While \eqref{eq:foot dynamics} governs the stance-foot evolution during continuous phases, the stance-foot pose undergoes a discrete jump at foot-landing events. At a contact switch occurring at $t=t_K^-$ ($K \in \mathbb{N}_+$), the change in the stance-foot orientation and position are computed using the forward-kinematics chain between the outgoing and incoming stance feet:
\begin{equation}
\label{eq:manual switch 1}
\small
\begin{aligned}
     \mathbf{R}^{F}_{K^+} &= \mathbf{R}^{F}_{K^-} \mathbf{h}_{R_{F,\text{old}}}^{\transpose}(\tilde{\mathbf{q}}_{K^-})\mathbf{h}_{R_{F,\text{new}}}(\tilde{\mathbf{q}}_{K^+}) + \mathbf{R}^{F}_{K^-} \mathbf{n}_{R_F} \\
    \mathbf{d}_{K^+} &= \mathbf{p}_{K^-} + \mathbf{R}^{F}_{K^-}  \mathbf{h}_{R_{F,\text{old}}}^{\transpose}(\tilde{\mathbf{q}}_{K^-}) \mathbf{h}_{p_{F,\text{new}}}(\tilde{\mathbf{q}}_{K^+}) + \mathbf{R}^{F}_{K^-} \mathbf{n}_{d} , 
\end{aligned}
\end{equation}
where {\small $(\cdot)_{K^-}$} and {\small $(\cdot)_{K^+}$} denote the values of $(\cdot)_t$ immediately before and after the switch, respectively.
Here, {\small $\mathbf{h}_{R_F} (\mathbf{q}_t) = {}^B\mathbf{R}^F_t$} and {\small $\mathbf{h}_{p_F}(\mathbf{q}) = {}^B\mathbf{p}^{BF}_t$} are the forward-kinematics mappings of foot orientation and position relative to the base, and the subscripts “old” and “new” indicate the previous and incoming stance feet. 
$\mathbf{n}_{R_F}$ and $\mathbf{n}_{d}$ capture additive uncertainty due to encoder noise, expressed as: 
{\small $\mathbf{n}_{R_F} :=- \mathbf{h}_{R_{F,\text{old}}}^{\transpose}(\tilde{\mathbf{q}}_{K^-})\mathbf{J}_{h_{R_{F,\text{new}}}}(\tilde{\mathbf{q}}_{K^+},\mathbf{w}^q_{K^+})
- \mathbf{J}_{h_{R_{F,\text{old}}}}^\transpose(\tilde{\mathbf{q}}_{K^-}, \mathbf{w}^q_{K^-}) \mathbf{h}_{R_{F,\text{new}}}(\tilde{\mathbf{q}}_{K^+})$}
and
{\small $ \mathbf{n}_{d} :=  -  \mathbf{h}_{R_{F,\text{old}}}^{\transpose}(\tilde{\mathbf{q}}_{K^-}) 
$} {\small $
\mathbf{J}_{h_{p_{F,\text{new}}}}(\tilde{\mathbf{q}}_{K^+})
\mathbf{w}^q_{K^+}
-  \mathbf{J}^\transpose_{h_{R_{F,\text{old}}}}(\tilde{\mathbf{q}}_{K^-}, \mathbf{w}^q_{K^-})  \mathbf{h}_{p_{F,\text{new}}}(\tilde{\mathbf{q}}_{K^+})$}.

Since the robot’s base pose and velocity relative to the dynamic ground do not change across foot-switch events, {\small $\mathbf{R}^B_t$, $\mathbf{p}_t$}, and {\small $\mathbf{v}_t$} remain continuous from {\small $t=t_K^-$} to {\small $t=t_K^+$}.

\vspace{-0.15 in}
\section{Filter Design} \label{sec:inekf}
\vspace{-0.05 in}
This section presents the design of the proposed InEKF based on the process and measurement models developed in Sec.~\ref{sec:model}, including the propagation and update steps.

\vspace{-0.15 in}
\subsection{Propagation Step during a Continuous Phase}
\vspace{-0.05 in}

During propagation, the InEKF predicts the estimate {\small $\bar{\mathbf{X}}_t$} and the error covariance {\small $\mathbf{P}_t$} using the process model. The corresponding measurement update, which incorporates sensor data to correct these predictions, is described later.

\subsubsection{State estimate propagation} 
The state estimate {\small $\bar{\mathbf{X}}_t$} is propagated using the deterministic component of the process models described in Sec.~\ref{sec:model}.

\subsubsection{Covariance estimate propagation} 

By the general InEKF methodology, the estimation-error covariance {\small $\mathbf{P}_t$} is propagated using the linearized log-error dynamics derived from the process model. Taking the time derivative of the right-invariant error in \eqref{eq:right-error} and applying the first-order approximations {\small $\boldsymbol{\eta}_t \approx \mathbf{I} + \boldsymbol{\xi}_t^{\wedge}$} and {\small $\tfrac{d}{dt}\boldsymbol{\eta}_t \approx \tfrac{d}{dt}(\boldsymbol{\xi}_t^{\wedge})$} yields:
\vspace{-0.05 in}
\begin{equation}
\vspace{-0.05 in}
\label{eq:linearized error}
\small 
    \tfrac{d}{dt} \boldsymbol{\xi}_t = \mathbf{A}_t \boldsymbol{\xi}_t + \mathbf{w}_{{A}_t},
\end{equation}
where {\small $\mathbf{A}_t$} is the Jacobian of the error dynamics and $\mathbf{w}_{{A}_t}$ is the process noise. Their expressions  are provided in the supplementary file.
Since \eqref{eq:linearized error} is linear, the covariance evolves according to the associated Riccati equation \cite{kalman1960new}: 
\vspace{-0.05 in}
\begin{equation}
\small
    \tfrac{d}{dt}(\mathbf{P}_t) = \mathbf{A}_t \mathbf{P}_t + \mathbf{P}_t \mathbf{A}^{\transpose}_t + \bar{\mathbf{Q}}_t,
    \vspace{-0.05 in}
\end{equation}
where {\small $\bar{\mathbf{Q}}_t =  {\mathbf{Q}}_{{A}_t}\text{Cov}(\mathbf{w}_{{A}_t}){\mathbf{Q}}_{{A}_t}^\transpose$} is constructed from the covariance of the IMU white noise {\small Cov(${}^i\mathbf{w}_t$)} for {\small $\{i \in {B,F}\}$}.

\vspace{-0.15 in}
\subsection{Measurement Update Step during a Continuous Phase}
\vspace{-0.05 in}

During the measurement update, the predicted state and covariance are corrected using the leg-odometry measurement model in \eqref{eq:FK} with newly available sensor data.

\subsubsection{State estimate update}

Since the measurement model admits a right-invariant observation form, the updated state estimate $\bar{\mathbf{X}}_t^{+}$ is given by \cite{barrau_invariant_2015}:
\vspace{-0.05 in}
\begin{equation}
\vspace{-0.05 in}
\small
    \bar{\mathbf{X}}^+_t = \exp(\mathbf{K}_t (\bar{\mathbf{X}}_t \mathbf{Y}_t - \mathbf{b}_t)) \bar{\mathbf{X}}_t,
\end{equation}
where $\mathbf{K}_{t}$ denotes the Kalman gain. 

\subsubsection{Covariance estimate update}

To update the covariance, we use the right-invariant InEKF property {\small $\mathbf{H}_t \boldsymbol{\xi}_t = -\boldsymbol{\xi}_t^\wedge \mathbf{b}_t$} \cite{barrau_invariant_2015}, and apply a first-order approximation of the right-invariant error, {\small $\boldsymbol{\eta}_t \approx \mathbf{I} + \boldsymbol{\xi}_t^\wedge$}. This yields the linearized update equation for the logarithmic error:
\vspace{-0.05 in}
\begin{equation}
\vspace{-0.05 in}
\small
\label{eq:xi+}
    \boldsymbol{\xi}_t^+ = \boldsymbol{\xi}_t - \mathbf{K}_t (\mathbf{H}_t \boldsymbol{\xi}_t - \bar{\mathbf{X}}_t \mathbf{n}_t),
\end{equation}
where the observation matrix $\mathbf{H}_t$ is: 
\vspace{-0.05 in}
\begin{equation}
\vspace{-0.05 in}
\small
    \mathbf{H}_t =
    \begin{bmatrix}
       \mathbf{0}_{3} &\mathbf{0}_{3} &-\mathbf{I}_{3}  &\mathbf{I}_{3} &\mathbf{0}_{3}
    \end{bmatrix} \label{eq:H}.
\end{equation}
Derivation of \eqref{eq:xi+} and \eqref{eq:H} is given in the supplementary file.

Since \eqref{eq:xi+} is linear, the Kalman gain follows from standard Kalman filtering as: {\small $\mathbf{K}_t =  \mathbf{P}_t {\mathbf{H}}_t^\transpose \mathbf{S}_t^{-1}$},
where {\small $ \mathbf{S}_t  =  \mathbf{H}_t {\mathbf{P}}_t \mathbf{H}_t^\transpose + \bar{\mathbf{N}}_t$}.
Here the measurement noise covariance {\small $\mathbf{\bar N}_t$} is given by:
{\small $\mathbf{\bar N}_t = \bar{\mathbf{N}}_R  \text{Cov}( \mathbf{w}_{t}^{q} ) \bar{\mathbf{N}}_R  ^\transpose $}, 
with 
{\small $\bar{\mathbf{N}}_R := \bar{\mathbf{R}}^F_t \mathbf{h}_{R}^{\transpose}(\tilde{\mathbf{q}_t})
(\mathbf{J}_{p}(\tilde{\mathbf{q}}_t) + [\mathbf{h}_{p}(\tilde{\mathbf{q}_t})]_{\times}\mathbf{J}_{\theta}(\tilde{\mathbf{q}_t}))$}, with derivations in the supplementary file. 

Finally,  the updated covariance estimate is:
\vspace{-0.05 in}
\begin{equation}
\vspace{-0.05 in}
\small
    \mathbf{P}_t^+ = \left(\mathbf{I}_{15} - \mathbf{K}_t\mathbf{H}_t \right) \mathbf{P}_t \left(\mathbf{I}_{15} - \mathbf{K}_t\mathbf{H}_t \right)^\transpose + \mathbf{K}_t \bar{\mathbf{N}}_t \mathbf{K}_t^\transpose.
\end{equation}

\vspace{-0.15 in}
\subsection{Propagation Step at a Discrete Contact Switch}
\vspace{-0.05 in}

\subsubsection{State estimate propagation}

At a discrete foot-landing event occurring at {\small $t=t_K^-$} ({\small $K \in \mathbb{N}+$}), the state undergoes a jump due to the change of stance foot. This discrete propagation can be approximated as:
\begin{equation}
\small
\begin{aligned}
    \mathbf{X}_K^+ &= \mathbf{X}_K^- 
    \begin{bmatrix}
        \mathbf{r}_1 & \0[3,1] & \0[3,1] & \mathbf{r}_2 & \0[3] \\
        \0[1,3] & 1 & 0 & 0 & \0[1,3] \\
        \0[1,3] & 0 & 1 & 1 & \0[1,3] \\
        \0[1,3] & 0 & 0 & 0 & \0[1,3] \\
        \0[3] & \0[3,1] & \0[3,1] & \0[3,1] & \mathbf{I}_3
    \end{bmatrix} - 
    \mathbf{X}_K^- 
    \begin{bmatrix}
        \mathbf{n}_{R_F} \\ \0[3,1] \\ \0[3,1] \\ \mathbf{n}_d \\ \0[3] 
    \end{bmatrix}^\wedge \\
    &=: \Delta_{u_t} (\mathbf{X}_{K^-},\tilde{\mathbf{q}}_{K^-}) - \mathbf{X}_{K^-} \mathbf{w}^\Delta_K ,
\end{aligned}
\end{equation} 
where
{\small $\mathbf{r}_1  := \mathbf{h}_{R_{F,\text{old}}}^{\transpose}(\tilde{\mathbf{q}}_{K^-})  \mathbf{h}_{R_{F,\text{new}}}(\tilde{\mathbf{q}}_{K^+}) $} and 
{\small $\mathbf{r}_2 := \mathbf{h}_{R_{F,\text{old}}}^{\transpose}(\tilde{\mathbf{q}}_{K^-}) $} {\small $\mathbf{h}_{p_{F,\text{new}}}(\tilde{\mathbf{q}}_{K^+}) $}. 
Thus,
the state can be propagated deterministically through {\small $ \Delta_{u_t} (\mathbf{X}_{K^-},\tilde{\mathbf{q}}_{K^-})$}.

\subsubsection{Covariance estimate propagation}

At a contact switch, the covariance is propagated through a linear transformation that removes the previous stance-foot state and introduces the new contact point. Accounting for foot impact and slippage, the covariance update follows \cite{hartley_contact-aided_2019}:
{\small ${\mathbf{P}}_t = {\mathbf{F}}_t {\mathbf{P}}_t {\mathbf{F}}_t^\transpose + {\mathbf{\bar{Q}}}_t^c $}, 
with derivations of {\small $\mathbf{F}_t$} and {\small ${\mathbf{\bar{Q}}}_t^c$} given in the supplementary file.

\begin{table} [t]
    \begin{center}
    \caption{Observability analysis.}
    \vspace{-0.05 in}
    \label{tab:obsv_analysis}
        \scalebox{0.85}
        {
        \begin{tabular}{|c|c|c|}
            \hline
            \multirow{2}{2em}{\textbf{Obs. type}} & \multirow{2}{8em}{\textbf{Foot-IMU reading}} &  \textbf{Observable} \\
            & & \textbf{pos. and vel.}\\
            \hline
            \textbf{\texttt{O1}} & ${}^F\tilde{\omega}_x = {}^F\tilde{\omega}_y = {}^F\tilde{\omega}_z = 0$ $\forall$ $t$ & $ \mathbf{v}_{t}$\\
            \hline
            \textbf{\texttt{\texttt{O2\!.\!1}}} & ${}^F\tilde{\omega}_x \not\equiv 0,$ ${}^F\tilde{\omega}_y = 0$, ${}^F\tilde{\omega}_z = 0$
            $\forall$ $t$ 
            & $\mathbf{v}_{t}$, ${p}_{y}$, ${p}_{z}$ \\ 
            \hline
            \textbf{\texttt{O2\!.\!2}} & ${}^F\tilde{\omega}_x = 0$, ${}^F\tilde{\omega}_y \not\equiv  0$, ${}^F\tilde{\omega}_z = 0$
            $\forall$ $t$
            & $\mathbf{v}_{t}$, ${p}_{x}$, ${p}_{z}$ \\ 
            \hline
            \textbf{\texttt{O2\!.\!3}} & ${}^F\tilde{\omega}_x = 0,$ ${}^F\tilde{\omega}_y = 0$, ${}^F\tilde{\omega}_z \not\equiv  0$ 
            $\forall$ $t$
            & $\mathbf{v}_{t}$,  ${p}_{x}$, ${p}_{y}$ \\ 
             \hline
            \textbf{\texttt{O3\!.\!1}} & ${}^F\tilde{\omega}_x \not\equiv  0$, ${}^F\tilde{\omega}_y \not\equiv  0$, ${}^F\tilde{\omega}_z = 0$ 
            $\forall$ $t$
            & \multirow{4}{3em}{$\mathbf{v}_{t}$, $\mathbf{p}_{t}$} \\
            \cline{1-2}
            \textbf{\texttt{O3\!.\!2}} & ${}^F\tilde{\omega}_x \not\equiv  0$, ${}^F\tilde{\omega}_y = 0$, ${}^F\tilde{\omega}_z \not\equiv  0$ 
            $\forall$ $t$
            & \\
            \cline{1-2}
            \textbf{\texttt{O3\!.\!3}} & ${}^F\tilde{\omega}_x = 0,$ ${}^F\tilde{\omega}_y \not\equiv  0$, ${}^F\tilde{\omega}_z \not\equiv  0$
            $\forall$ $t$
            & \\
            \cline{1-2}
            \textbf{\texttt{O3\!.\!4}} & ${}^F\tilde{\omega}_x \not\equiv  0,$ ${}^F\tilde{\omega}_y \not\equiv  0$, ${}^F\tilde{\omega}_z \not\equiv  0$ 
            $\forall$ $t$
            & \\
            \hline
        \end{tabular}
        }
    \end{center}
    \vspace{-0.25 in}
\end{table}
\vspace{-0.15 in}
\section{Observability Analysis} \label{sec:obsv}
\vspace{-0.1 in}


\subsubsection{Observability matrix}

We perform a local observability analysis on the linearized log-error dynamics in \eqref{eq:linearized error}. Since the filter operates in discrete time with discrete measurements, a discrete-time observability analysis is adopted.

The local observability between adjacent measurement update steps (i.e., over {\small [$t_k$, $t_{k+1}$]} with {\small $k \in \mathbb{N}_+$}) depends on both the state-transition matrix $\boldsymbol{\Phi}(t_k, t_{k+1})$, and the output matrix {\small $\mathbf{H}_t$} in \eqref{eq:H}.
The state-transition matrix satisfies 
{\small $\Dot{\boldsymbol{\Phi}}(t_k, t_{k+1})  = \mathbf{A}_k\boldsymbol{\Phi}(t_k, t_{k+1})$}
with {\small $\boldsymbol{\Phi}(t_k, t_k) = \mathbf{I}_{15}$}, 
and admits the closed-form solution
{\small $\boldsymbol{\Phi}(t_k, t_{k+1}) = \exp_m(\mathbf{A}_k \Delta{t})$}:
\begin{equation}
\small 
    \boldsymbol{\Phi}_k:=\boldsymbol{\Phi}(t_k, t_{k+1}) =
    \begin{bmatrix}
        \mathbf{I}_3 & \mathbf{0}_3 & \mathbf{0}_3 & \mathbf{0}_3 & \mathbf{0}_3 \\
        \mathbf{0}_3 & \boldsymbol{\Phi}_{k,22} & \mathbf{0}_3 & \boldsymbol{\Phi}_{k,24} & \mathbf{0}_3\\
        \mathbf{0}_3 & \boldsymbol{\Phi}_{k,32} & \boldsymbol{\Phi}_{k,33} & \boldsymbol{\Phi}_{k,34} & \mathbf{0}_3 \\
        \mathbf{0}_3 & \mathbf{0}_3 & \mathbf{0}_3 & \mathbf{I}_3 & \mathbf{0}_3 \\
        \mathbf{0}_3 & \mathbf{0}_3 & \mathbf{0}_3 & \mathbf{0}_3 & \mathbf{I}_3 \\
    \end{bmatrix},
\end{equation}
where the expressions of {\small $\boldsymbol{\Phi}_{k,22}$}, {\small $\boldsymbol{\Phi}_{k,24}$}, {\small $\boldsymbol{\Phi}_{k,32}$}, {\small $\boldsymbol{\Phi}_{k,33}$}, and {\small $\boldsymbol{\Phi}_{k,34}$} are derived in the supplementary file.

The discrete-time local observability matrix at {\small $\bar{\mathbf{X}}_t$} is \cite{chen1990local}:
\begin{small}
    \begin{equation}
    \vspace{-0.05 in}
        \boldsymbol{\mathcal{O}} =
        \begin{bmatrix}
            \mathbf{H}_k
            \\
            \mathbf{H}_{k+1}\boldsymbol{\Phi}_k
            \\
            \mathbf{H}_{k+1}\boldsymbol{\Phi}_k \boldsymbol{\Phi}_{k+1}
            \\
            \vdots
        \end{bmatrix}
        =
        \begin{bmatrix}
            \0[3] &\0[3] &-\I[3] &\I[3] &\0[3]
            \\
            \0[3] &\mathcal{O}_{22} &\mathcal{O}_{23} &\mathcal{O}_{24} &\0[3]
            \\
            \0[3] &\mathcal{O}_{32} &\mathcal{O}_{33} & \mathcal{O}_{34} &\0[3]
            \\
            \vdots &\vdots &\vdots &\vdots &\vdots
        \end{bmatrix},
    \end{equation}
\end{small}
\noindent where elements of $\boldsymbol{\mathcal{O}}$ are provided in the supplementary file.

\subsubsection{Observability of base position, velocity, and foot position}

From the expanded expression of $\boldsymbol{\mathcal{O}}$ in the supplementary file, the observability of the base position $\mathbf{p}_t$, base velocity $\mathbf{v}_t$, and stance-foot position $\mathbf{d}_t$ depends solely on the stance-foot gyroscope measurement
{\small ${}^{F}\tilde{\boldsymbol{{\omega}}}_{t}=:[{}^{F}\tilde{{\omega}}_{x},{}^{F}\tilde{{\omega}}_{y},{}^{F}\tilde{{\omega}}_{z}]^\transpose$}. 
Given the fast sampling of IMUs, the gyroscope readings can be assumed constant over {\small $[t_k,t_{k+1}]$}, enabling analytical observability characterization (see supplementary material).

As summarized in Tab.~\ref{tab:obsv_analysis}, the relative base velocity $\mathbf{v}_t$ is always observable, regardless of whether the ground is stationary or moving (observability type \textbf{\texttt{O1}}).
If only one principal axis of the stance-foot frame $\{F\}$ exhibits nontrivial angular-motion readings, the base position $\mathbf{p}_t=[p_x,p_y,p_z]^\transpose$ is observable except along that axis (types \textbf{\texttt{O2\!.\!1}}-\textbf{\texttt{O2\!.\!3}}).
When at least two axes of $\{F\}$ have nonzero angular-motion readings, the full base position becomes observable (types \textbf{\texttt{O3\!.\!1}}-\textbf{\texttt{O3\!.\!4}}). 

\subsubsection{Reconstruction of base orientation via base IMU augmentation}

The observability matrix {\small $\boldsymbol{\mathcal{O}}$} further indicates that the base orientation $\mathbf{R}^B_t$ and stance-foot orientation $\mathbf{R}^F_t$ are unobservable under the nominal sensing configuration. Here, the nominal sensing configuration refers to the proposed FRS filter using a single torso IMU together with the IMU on the current stance foot during a given stance phase. Yet, since the relative base position $\mathbf{p}_t$ is fully observable under general ground motions (Tab.~I), this property can be used to recover base-orientation observability through sensor augmentation. 

Specifically, we consider an augmented sensing configuration with three non-collinear IMUs rigidly mounted on the torso, so the filter can estimate the position of each IMU using the proposed formulation. These three position estimates define a plane fixed to the torso, which uniquely determines the base orientation, thereby enabling full-state observability through torso-IMU augmentation, rather than under the nominal sensing configuration alone. An example of this reconstruction is demonstrated experimentally in Sec.~VII.

\begin{figure}[t]
    \centering
    \includegraphics[width=0.8 \linewidth]{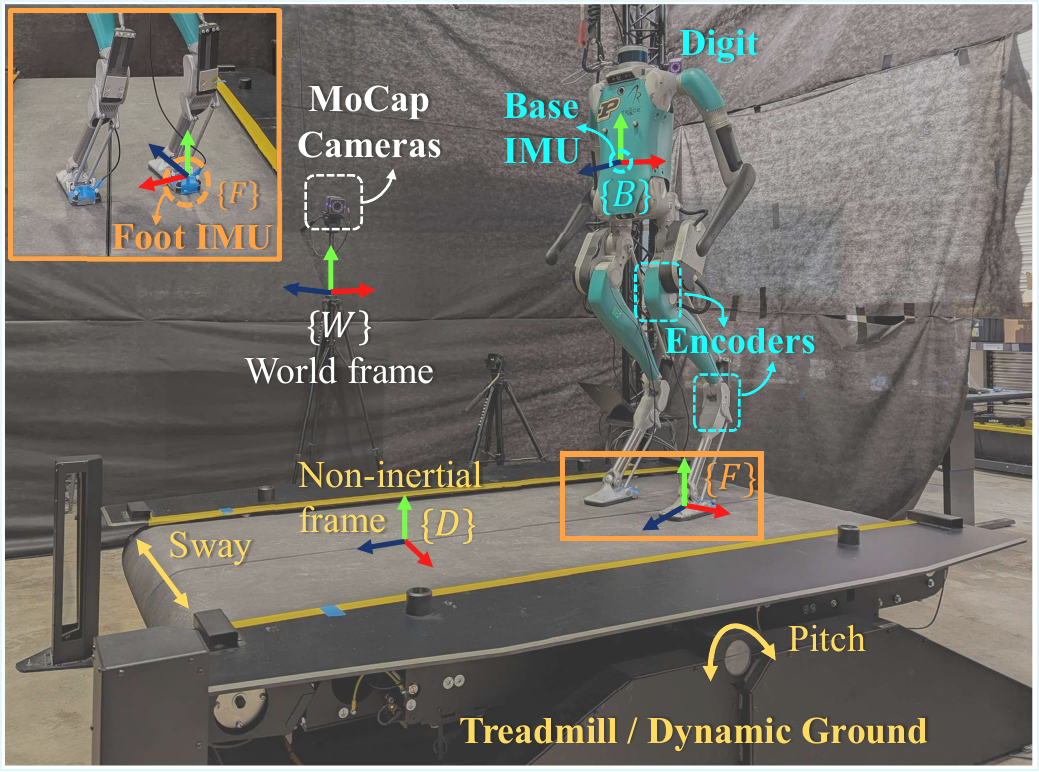}
    \vspace{-0.05 in}
    \caption{Experimental setup for performance validation on Digit using torso and foot IMUs, encoders, and motion-capture ground truth on a dynamic treadmill.}
    \label{fig:exp_setup}
    \vspace{-0.15 in}
\end{figure}

\vspace{-0.15 in}
\section{Experimental Validation} \label{sec:exp}
\vspace{-0.05 in}

This section reports the experimental setup and validation results, including comparison to baseline filters. 
An experiment movie is provided in the supplementary material and available at: \href{https://youtu.be/3AsKlwrB1ag}{https://youtu.be/3AsKlwrB1ag}.

\vspace{-0.15 in}
\subsection{Experimental Setup}
\vspace{-0.05 in}

\subsubsection{Robot and sensors}
Experiments are conducted on the Agility Robotics Digit humanoid robot. The proprietary torso-mounted IMU  is used as the base-frame IMU while onboard IMU and joint encoder data are sampled at 500~Hz. Two external IMUs (WT901C-232, WitMotion Co., Ltd.) are mounted on the footpads and record data at 200~Hz. All onboard sensors are time-synchronized through a single computation platform. The 200 Hz foot-end IMU measurements are then linearly interpolated to 500 Hz before filter fusion. The sensor specifications are summarized in the supplementary material. A Vicon mo-cap system provides ground-truth state values.

\begin{figure}[t]
    \centering
    \includegraphics[width= 0.9\linewidth]{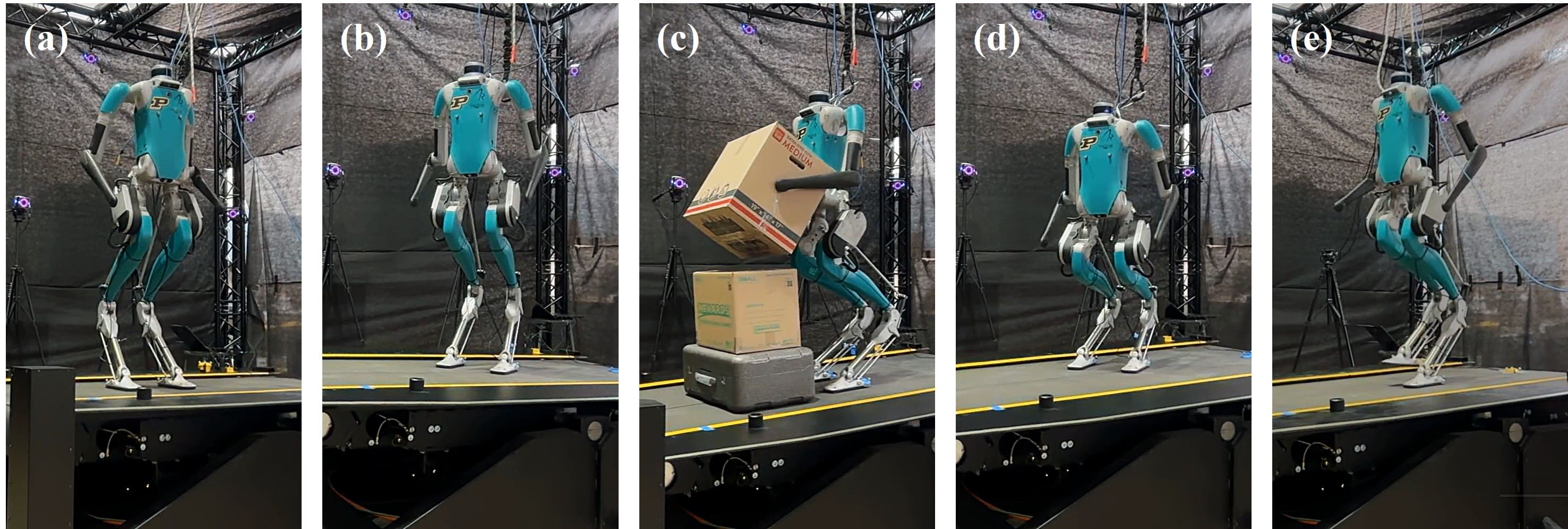}
    \vspace{-0.05 in}
    \caption{Snapshots of Digit operating on a moving treadmill under five representative test cases: (a) Case {\textbf{\texttt{T1}}}: standing with a $90\degree$ heading relative to the treadmill; (b) Case {\textbf{\texttt{T2}}}: standing with a $45\degree$ heading; (c) Case {\textbf{\texttt{T3}}}: box picking and placing; (d) Case {\textbf{\texttt{T4}}}: squatting; and (e) Case {\textbf{\texttt{T5}}}: walking.}
    \vspace{-0.25 in}
    \label{fig:placement}
\end{figure}

\subsubsection{Dynamic ground} Dynamic ground motion is emulated using a Motek M-Gait treadmill (Fig.~\ref{fig:exp_setup}) programmed with the following sway and pitch trajectories:
({\textbf{\texttt{GM1}}}) $4 \cos\left({\frac{2\pi}{6}}t\right)$~cm; 
({\textbf{\texttt{GM2}}}) $4 \cos\left({\frac{2\pi}{5}}t\right)$~cm; 
({\textbf{\texttt{GM3}}}) $\ang{4} \sin\left({\frac{2\pi}{6}}t\right)$; 
({\textbf{\texttt{GM4}}}) $\ang{4} \sin\left({\frac{2\pi}{8}}t\right)$; 
and
({\textbf{\texttt{GM5}}}) $\ang{4} \sin\left({\frac{2\pi}{7}}t\right)$. 

\subsubsection{Test cases}

Five experimental cases are designed to assess filter performance under diverse robot and ground motions: two standing trials, two squatting trials (with and without lifting a box), and one walking trial, each conducted under different ground motions. The test cases are illustrated in Fig.~\ref{fig:placement} and summarized in Tab.~\ref{tab:exp_cases}. Guided by the observability analysis, Digit is oriented with different headings, inducing distinct stance-foot angular motions and enabling evaluation under varying observability conditions.

\subsubsection{Baseline estimators}

The proposed estimator, termed ``FRS'', is compared against two InEKF-based baselines \cite{hartley_contact-aided_2019, he_invariant_2025}, both of which propagate the state using IMU motion dynamics and perform measurement updates via leg odometry.

The first baseline, ``SRS'' \cite{hartley_contact-aided_2019}, is designed for legged robots operating on static terrain. In the absence of IMU noise and bias, its process model satisfies the group-affine property and its measurement model is right-invariant. However, the static-ground assumption restricts its applicability to stationary environments and results in unobservable base yaw and position.

The second baseline, ``DRS'' \cite{he_invariant_2025}, estimates the robot’s pose and velocity relative to a dynamic ground frame by explicitly modeling ground motion. While the filter achieves full state observability, its main limitation is the reliance on an external ground-mounted IMU, which may be impractical in real-world deployments. 
Results from an additional baseline, DogLegs~\cite{wu2025doglegs}, are provided in the supplementary material.

\begin{figure*}
     \centering
     \begin{subfigure}[b]{0.48\textwidth}
         \centering
         \includegraphics[width=\textwidth,frame]{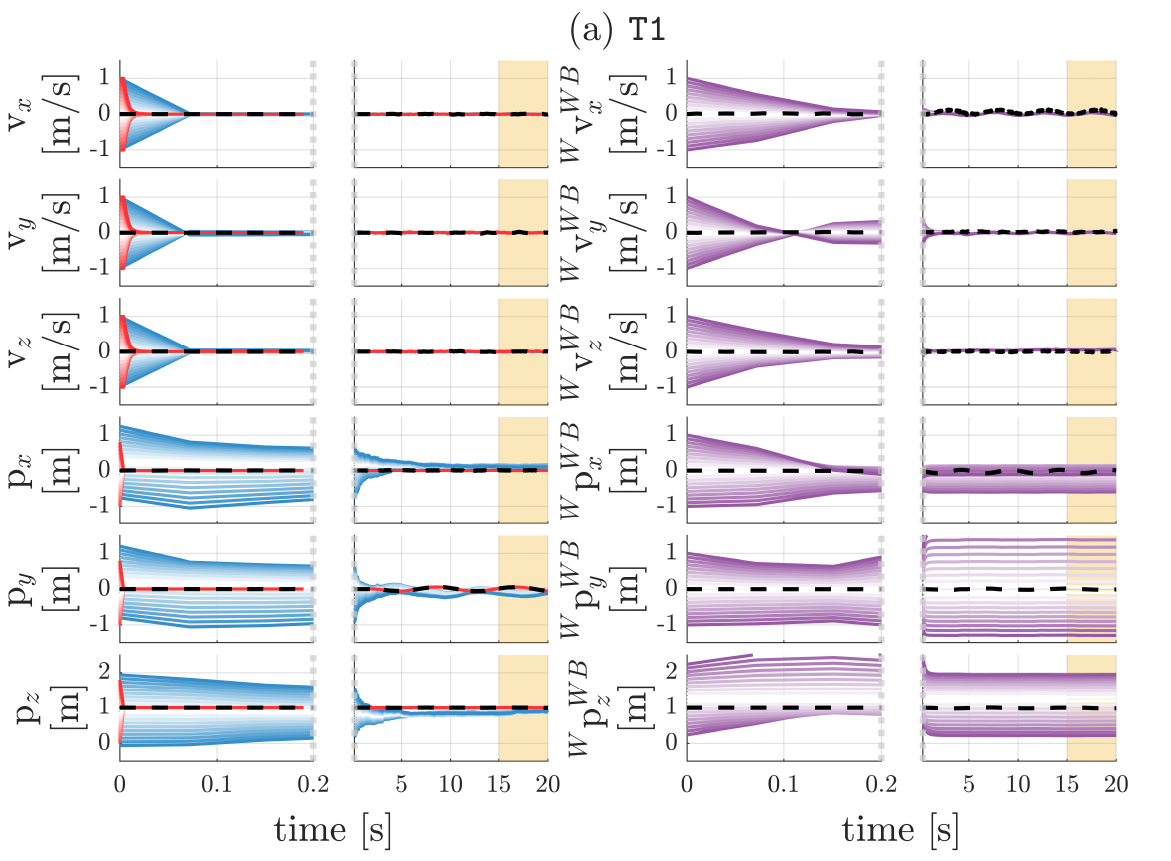}
     \end{subfigure}
     \hspace{-0.2cm}
     \begin{subfigure}[b]{0.48\textwidth}
         \centering
         \includegraphics[width=\textwidth,frame]{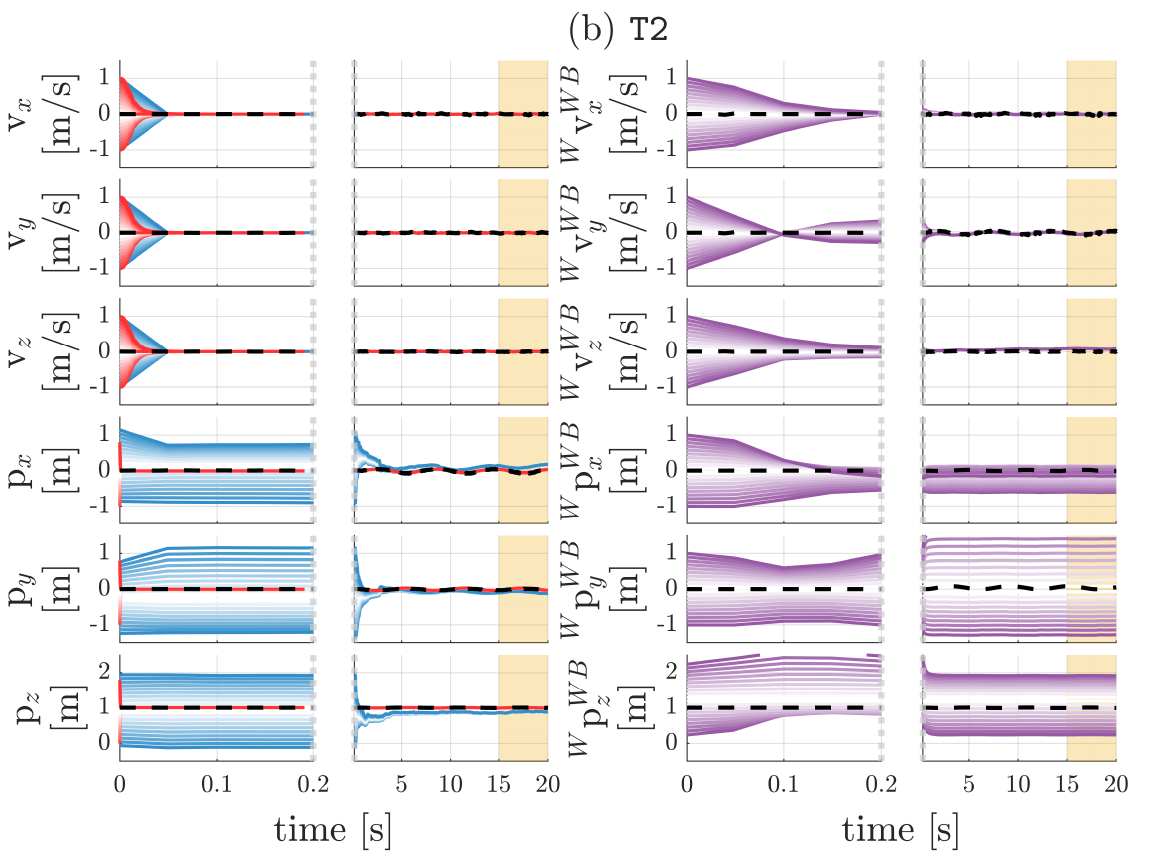}
     \end{subfigure}
     \begin{subfigure}[b]{0.48\textwidth}
         \centering
         \includegraphics[width=\textwidth, frame]{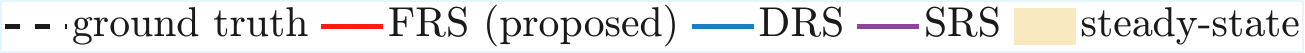}
         \label{fig:legend}
     \end{subfigure}
     \vspace{-0.2 in}
     \caption{Estimation results of the proposed FRS and the baseline DRS and SRS filters during Digit balancing on a swaying and pitching ground for test cases $\textbf{\texttt{T1}}$ and $\textbf{\texttt{T2}}$ under significant initial errors.
    For FRS and DRS, the base velocity $\mathbf{v}=:[v_x,v_y,v_z]^\transpose$ and position $\mathbf{p}=:[p_x,p_y,p_z]^\transpose$ relative to the dynamic ground are shown. For SRS, the absolute base velocity ${}^W \mathbf{v}^{WB}=:[{}^W{v}^{WB}_x,{}^W{v}^{WB}_y,{}^W{v}^{WB}_z]^\transpose$ and position ${}^W \mathbf{p}^{WB}=:[{}^W{p}^{WB}_x,{}^W{p}^{WB}_y,{}^W{p}^{WB}_z]^\transpose$ in the inertial world frame are shown.}
    \label{fig:compare_T1_T2}
        \vspace{-0.05 in}
\end{figure*}
\begin{figure*}
     \centering
     \begin{subfigure}[b]{0.48\textwidth}
         \centering
         \includegraphics[width=\textwidth,frame]{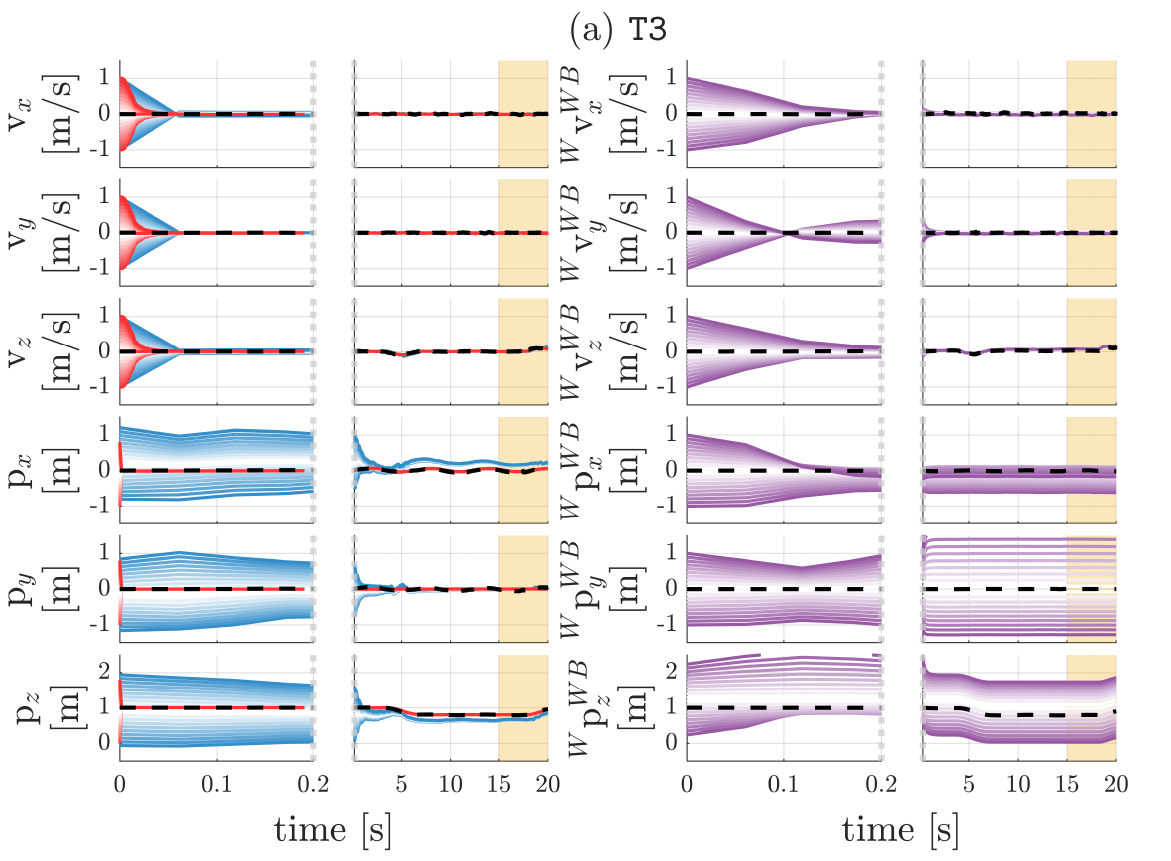}
     \end{subfigure}
     \hspace{-0.2cm}
     \begin{subfigure}[b]{0.48\textwidth}
         \centering
         \includegraphics[width=\textwidth,frame]{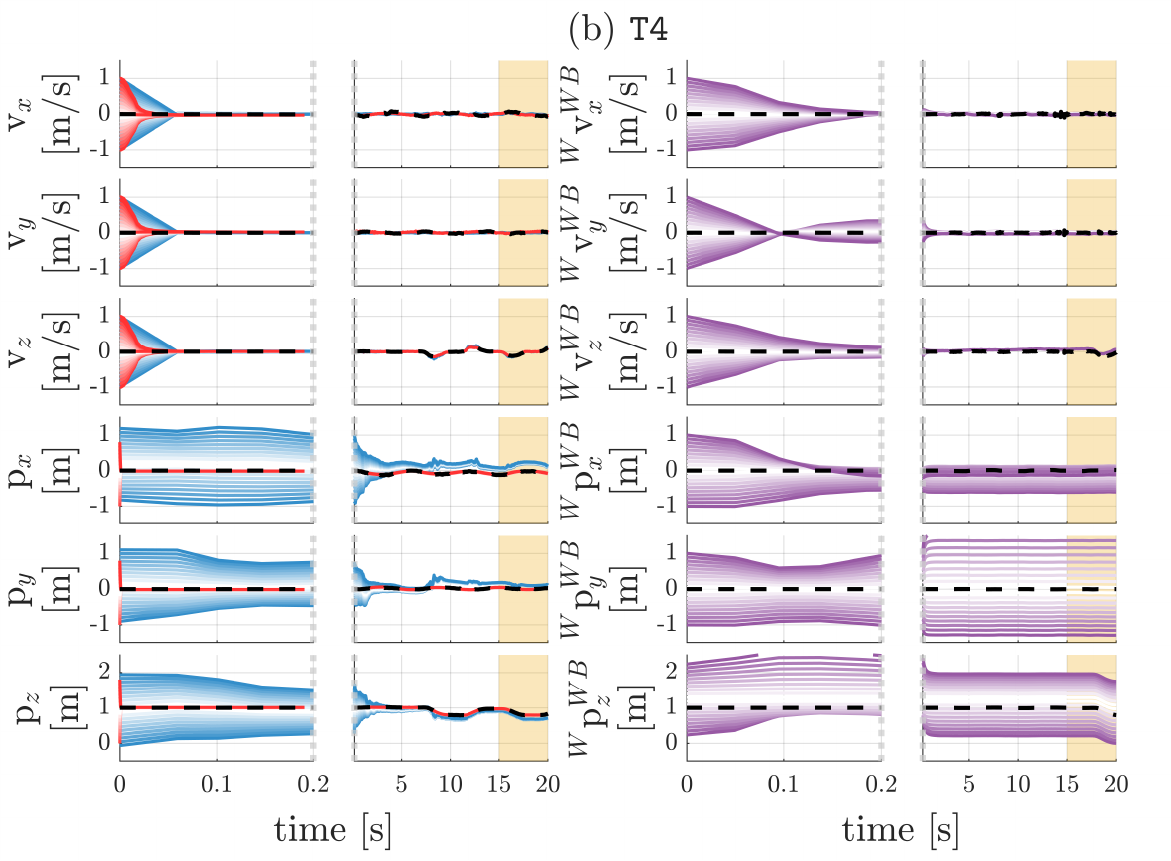}
     \end{subfigure}
     \begin{subfigure}[b]{0.48\textwidth}
         \centering
         \includegraphics[width=\textwidth,frame]{figures/legend.pdf}
         \label{fig:legend}
     \end{subfigure}
     \vspace{-0.2 in}
    \caption{Estimation results of the proposed FRS and the baseline DRS and SRS filters during Digit squatting on a pitching ground (i.e., under test cases $\textbf{\texttt{T3}}$ and $\textbf{\texttt{T4}}$) under significant initial errors. The notational convention follows Fig.~\ref{fig:compare_T1_T2}.}
    \vspace{-0.2 in}
    \label{fig:compare_T3_T4}
\end{figure*}

For fair comparison under large initial errors, all filters are evaluated on the same experimental data, with initial base velocity and position errors uniformly sampled from $[-1,1]$~m/s and $[-1,1]$~m, respectively. Covariance matrices are tuned based on nominal sensor noise characteristics and refined empirically to achieve best performance. The estimation results are shown in Figs.~\ref{fig:compare_T1_T2}, \ref{fig:compare_T3_T4}, and \ref{fig:compare_T5} and summarized in Tab.~\ref{tab:accuracy}, where the baseline performance is consistent with previously reported results \cite{he_invariant_2025}.

\begin{small}
\begin{table}
    \begin{center}
    \caption{Summary of the test cases under various robot and ground motions and different observability types.}
    \vspace{-0.05 in}
    \label{tab:exp_cases}
    \scalebox{0.85}
    {
        \begin{tabular}{|c|c|c|c|c|} 
            \hline
            \multirow{2}{2em}{\textbf{Test case}}  & \textbf{Digit} & \textbf{Digit} & \textbf{Ground} & \textbf{Predicted} \\
            & \textbf{motion} & \textbf{heading} & \textbf{motion} & \textbf{obs. type} \\
            \hline
             {\textbf{\texttt{T1}}} & {Standing} & \ang{90} & {\textbf{\texttt{GM2}}}\,+\,{\textbf{\texttt{GM4}}} & \textbf{\texttt{O2\!.\!1}} \\
             \hline
            {\textbf{\texttt{T2}}} &  {Standing} & \ang{45} & {\textbf{\texttt{GM1}}}\,+\,{\textbf{\texttt{GM3}}}  & \textbf{\texttt{O3\!.\!1}} \\
            \hline
            {\textbf{\texttt{T3}}} &  {Squatting} & \ang{0} & {\textbf{\texttt{GM3}}}  & \textbf{\texttt{O2\!.\!2}}\\
            \hline
            {\textbf{\texttt{T4}}} & {Squatting} & \ang{45} & {\textbf{\texttt{GM3}}} & \textbf{\texttt{O3\!.\!1}} \\
            \hline
            {\textbf{\texttt{T5}}} & Walking & \ang{0} & {\textbf{\texttt{GM5}}} & \textbf{\texttt{O2\!.\!2}} \\
            \hline
        \end{tabular}}
    \end{center}
    \vspace{-0.2 in}
\end{table}
\end{small}

\small
\begin{table}
    \begin{center}
    \caption{Maximum steady-state absolute position error for FRS/DRS under varying initial errors. The steady-state periods are as specified in Figs.~\ref{fig:compare_T1_T2}, \ref{fig:compare_T3_T4}, and \ref{fig:compare_T5}.}
    \vspace{-0.07 in}
    \label{tab:accuracy}
    \scalebox{0.88}
    {
        \begin{tabular}{|c|c|c|c|}
            \hline
            {\textbf{Test case}} & $p_x$\,[m] & $p_y$\,[m] & $p_z$\,[m]\\
            \hline
            {\textbf{\texttt{T1}}} & \underline{0.007} / 0.148 & \underline{0.006} / 0.199 & \underline{0.003} / 0.170 \\
            \hline
            {\textbf{\texttt{T2}}} & \underline{0.011} / 0.138 & \underline{0.012} / 0.087 & \underline{0.009} / 0.128 \\
            \hline
            {\textbf{\texttt{T3}}} & \underline{0.018} / 0.194 & 0.039 / \underline{0.029} & \underline{0.013} / 0.149 \\
            \hline
            {\textbf{\texttt{T4}}} & \underline{0.013} / 0.240 & \underline{0.011} / 0.112 & \underline{0.007} / 0.115 \\
            \hline
            {\textbf{\texttt{T5}}} & \underline{0.113} / 0.136 & \underline{0.067} / 0.115 & 0.057 / \underline{0.025} \\
            \hline
        \end{tabular}}
    \end{center}
    \vspace{-0.25 in}
\end{table}
\normalsize

\vspace{-0.1 in}
\subsection{Results Analysis and Comparison}
\vspace{-0.05 in}

\subsubsection{Observability of base position and velocity} 

Based on the theoretical observability analysis in Tab.~\ref{tab:obsv_analysis} and the predicted observability for each test case in Tab.~\ref{tab:exp_cases}, the relative base velocity is expected to be observable in all experiments. The relative base position is predicted to be partially observable in Cases \textbf{\texttt{T1}} ($p_x$ unobservable) and \textbf{\texttt{T3}} and \textbf{\texttt{T5}} ($p_y$ unobservable), and fully observable in Cases \textbf{\texttt{T2}} and \textbf{\texttt{T4}}.

Experimentally, a state is considered observable if its estimate converges sufficiently close to the ground truth \cite{hartley_contact-aided_2019}. Under this criterion, the theoretical predictions are largely confirmed by the results in Figs.~\ref{fig:compare_T1_T2}a, \ref{fig:compare_T3_T4}a, and \ref{fig:compare_T5}. The only discrepancies occur for $p_x$ in Case \textbf{\texttt{T1}}, and $p_y$ in Cases \textbf{\texttt{T3}} and \textbf{\texttt{T5}}.

Notably, these position components also converge to small neighborhoods around the ground truth, indicating practical observability under experimental conditions. This behavior is likely due to non-ideal experimental conditions (e.g., slight treadmill motion errors and small misalignment between Digit and the treadmill), which induce additional foot IMU excitation and effectively restore full positional observability. 
Yet, this does not alter the theoretical observability result or provide a formal guarantee. Thus, when full position observability is required, deployment should avoid this degenerate single-axis condition rather than rely on incidental non-idealities.

In contrast, the base position estimates under the baseline SRS filter fail to converge, consistent with its known observability limitations \cite{hartley_contact-aided_2019}. The baseline DRS filter exhibits observability behavior similar to FRS, confirming previous observability results~\cite{hartley_contact-aided_2019}.

\subsubsection{Real-time computation} 
Owing to their closed-form formulations, all three filters support real-time implementation, with average computation times of less than 3 ms per cycle on a laptop with a 13th-Gen Intel Core i7-13620H processor and 16 GB RAM.

\subsubsection{Error convergence rate}
For base velocity estimation, all three filters ensure convergence to a small neighborhood of the ground truth within approximately 1\,sec for all test cases, with the proposed FRS filter being consistently the fastest.
With respect to base position estimation across all cases, the proposed FRS converges within 0.1\,sec whereas the baseline DRS takes 3-5\,sec on average.

The significantly faster convergence of FRS is likely due to its measurement model incorporating both positional and rotational forward kinematics between the base and the foot, whereas DRS and SRS rely only on positional kinematics.

\subsubsection{Steady-state accuracy}

\begin{figure}
     \centering
     \begin{subfigure}[b]{0.48\textwidth}
         \centering
         \includegraphics[width=\textwidth,frame]{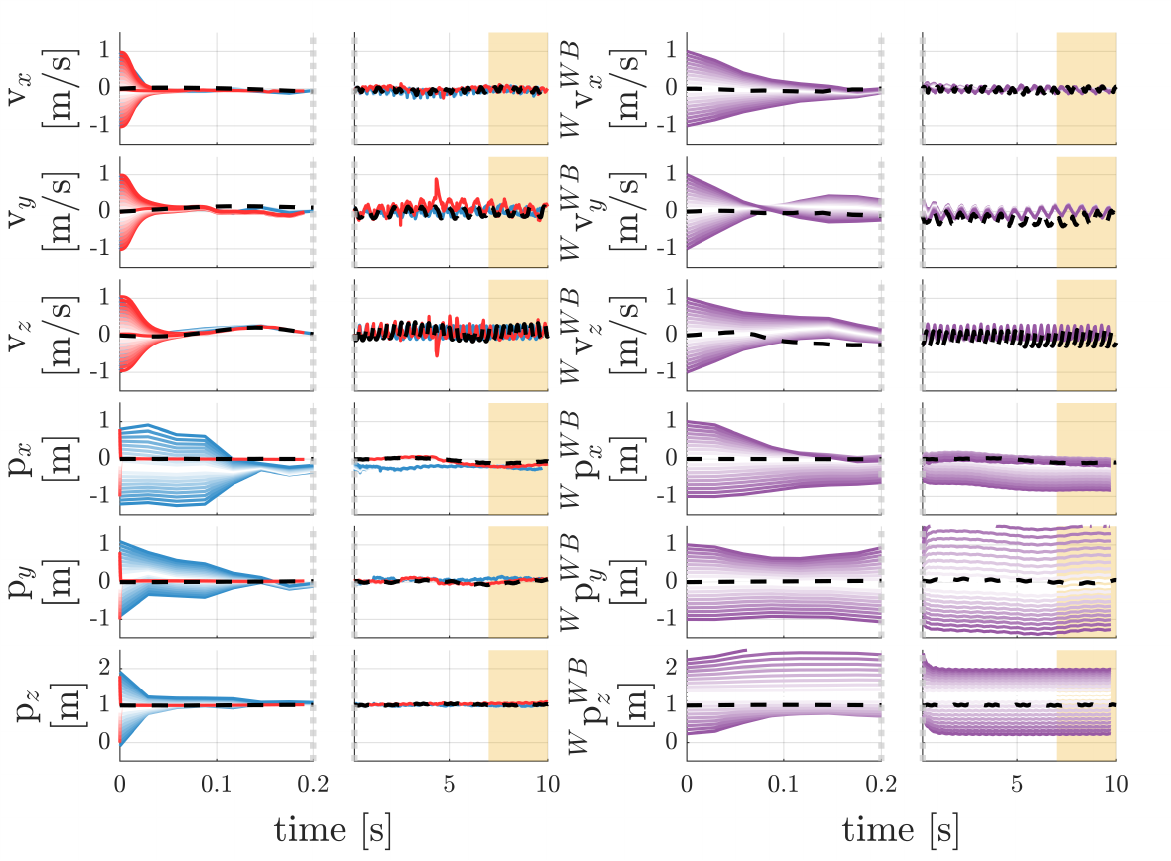}
     \end{subfigure}
     \begin{subfigure}[b]{0.43\textwidth}
         \centering
         \includegraphics[width=\textwidth,frame]{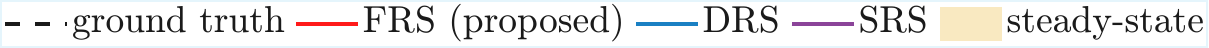}
     \end{subfigure}
    \caption{Estimation results of the proposed FRS and the baseline DRS and SRS filters during Digit walking on a pitching ground for test case $\textbf{\texttt{T5}}$ under significant initial errors. The notational conversion follows Fig.~\ref{fig:compare_T1_T2}.}
    \label{fig:compare_T5}
    \vspace{-0.2 in}
\end{figure}

As shown in Figs.~\ref{fig:compare_T1_T2} and \ref{fig:compare_T3_T4}, all three filters achieve comparable steady-state accuracy in base velocity estimation for Cases \textbf{\texttt{T1}}-\textbf{\texttt{T4}}. In Case \textbf{\texttt{T5}} (Fig.~\ref{fig:compare_T5}), the proposed FRS exhibits noticeably larger steady-state velocity errors than in the standing and squatting cases. This degradation is primarily attributed to impact-induced noise in the foot-mounted IMUs during walking, an effect further amplified by the moving ground.

For base position estimation, the proposed FRS consistently outperforms the SRS baseline and achieves the best overall performance among the three filters in Cases \textbf{\texttt{T1}}-\textbf{\texttt{T4}}, in terms of both transient response and steady-state accuracy. As discussed earlier, SRS fails to ensure position observability on non-inertial ground and therefore does not converge, resulting in significantly larger position errors. In contrast, both FRS and DRS render the base position observable.

As summarized in Tab.~\ref{tab:accuracy}, FRS achieves steady-state position errors below 4\,cm in Cases \textbf{\texttt{T1}}-\textbf{\texttt{T4}}, compared to approximately 10\,cm for DRS. In Case \textbf{\texttt{T5}}, FRS maintains good performance through contact-noise mitigation, including mechanical impact absorption at the feet, 
stable contact switching during double support,
and low-pass filtering of IMU data. These measures are detailed in the supplementary file, which also compares the results with an unmitigated experiment. As a result, FRS outperforms SRS and, despite residual jumps in the velocity estimates, achieves better $x$- and $y$-position estimates than DRS using only onboard sensing.

\subsubsection{Reconstruction of base orientation with filter augmentation} \label{sec:D_R_B}

\begin{figure}[t]
    \centering
    \includegraphics[width=\linewidth]{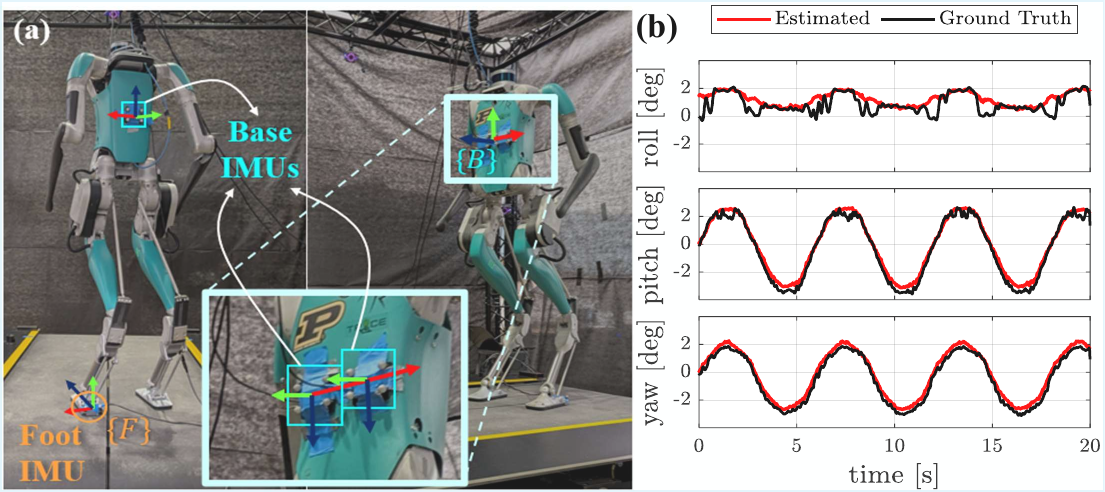}
    \vspace{-0.15in}
    \caption{(a) Experimental setup for base-orientation reconstruction using three torso-mounted IMUs under the FRS framework. (b) Reconstructed relative base orientation $\mathbf{R}^B_t$ compared with ground truth, validating accurate orientation recovery through IMU augmentation.}
    \label{fig:orientation est setup}
    \vspace{-0.3 in}
\end{figure}

To validate the proposed IMU augmentation strategy for recovering base-orientation observability, three IMUs are mounted at distinct, non-collinear locations on the robot torso, as illustrated in Fig.~\ref{fig:orientation est setup}. These IMU positions define a plane rigidly attached to the torso, from which an effective base frame $\{B\}$ can be reconstructed.

Experimental data are collected while Digit stands on the treadmill undergoing pitching motion \textbf{\texttt{GM3}}, with the robot oriented at $\ang{15}$. Under this test condition, the relative base position is fully observable, enabling reconstruction of the base orientation as described in Sec.~\ref{sec:obsv}. The reconstructed base orientation closely matches the motion-capture-based ground truth, as shown in Fig.~\ref{fig:orientation est setup}. The mean absolute errors are 0.3904°, 0.2709°, 0.3241° in roll, pitch, and yaw, respectively.
The small residual discrepancies (below $\ang{2}$) are primarily attributed to curved mounting surfaces on the torso and limited precision in relative IMU placement. Still, the results confirm that augmenting the proposed FRS framework with multiple torso-mounted IMUs enables reliable recovery of base orientation, thereby achieving full state observability in practice.

\vspace{-0.15 in}
\section{Conclusion} \label{sec:conclusion}
\vspace{-0.05 in}

This paper has presented a proprioceptive invariant extended Kalman filter for state estimation of humanoid robots operating on accelerating ground. By exploiting foot-mounted inertial measurement units (IMUs), the filter explicitly accounts for ground-induced nonlinearities without requiring direct ground-motion measurements or external instrumentation.
Observability analysis showed that the robot’s base position and velocity relative to the moving ground are observable under general ground motions and that base orientation can be recovered in practice through simple torso-IMU augmentation.
Experiments on the Digit humanoid robot during standing, squatting, and walking on a moving treadmill validated the proposed approach and demonstrated fast convergence, high accuracy, and robustness to impact-induced foot-IMU noise, even under large initial errors. Compared with existing InEKF baselines for static and instrumented dynamic ground, the proposed filter achieves comparable or superior performance using solely onboard sensing. Our future work will explore the integration of the proposed filter with locomotion planners and controllers formulated directly in the dynamic ground frame.

\vspace{-0.15 in}
\bibliographystyle{IEEEtran} 
\bibliography{reference_new}

\end{document}

%% file: definition.tex
\usepackage{graphicx}
\usepackage{caption,booktabs}

\graphicspath{
    {figures/}
}

\usepackage{amsmath,amssymb,enumerate}

\usepackage{amsthm}
\usepackage{setspace}
\usepackage{booktabs}
\usepackage[dvipsnames,svgnames,table]{xcolor}
\usepackage{tabularx,ragged2e}
\usepackage{mathtools}
\usepackage{algorithm, algorithmicx, algpseudocode}

\usepackage{blindtext}
\usepackage{gensymb}
\usepackage{framed}
\usepackage{xparse}
\usepackage{lipsum}
\usepackage{makecell}
\usepackage{subcaption}
\usepackage{soul} 
\usepackage{mathrsfs}
\usepackage[mathscr]{euscript}
\usepackage{times}
\usepackage{cite} 
\usepackage{multicol}
\usepackage{multirow}
\usepackage{diagbox}
\definecolor{darkgreen}{rgb}{0,0.6,0}
\definecolor{darkred}{rgb}{0.6,0,0}
\usepackage[bookmarks=true,colorlinks=true,pdfpagemode=UseNone,citecolor=darkgreen,linkcolor=darkred,urlcolor=blue]{hyperref}

\usepackage{amsfonts}
\usepackage[utf8]{inputenc}
\usepackage[T1]{fontenc}
\usepackage{textcomp}
\usepackage{arydshln}
\usepackage[normalem]{ulem}
\usepackage{siunitx}
\usepackage[export]{adjustbox}
\usepackage[dvipsnames]{xcolor}
\usepackage{cancel}

\renewcommand{\thefigure}{\arabic{figure}}

\definecolor{note}{rgb}{0.1,0.1,1}
\definecolor{rephase}{rgb}{0.15,0.7,0.15}
\definecolor{bag}{rgb}{0.6,0.6,0.2}

\makeatletter
\renewcommand*\env@matrix[1][c]{\hskip -\arraycolsep
  \let\@ifnextchar\new@ifnextchar
  \array{*\c@MaxMatrixCols #1}}
\makeatother

\newcommand{\transpose}{\mathsf{T}}

\newcolumntype{Y}{>{\RaggedRight\arraybackslash}X}

\makeatletter
\newcommand{\mathleft}{\@fleqntrue\@mathmargin0pt}
\newcommand{\mathcenter}{\@fleqnfalse}
\makeatother

\newcommand{\Rt}[1][]{\mathbf{R}^{#1}_t}

\newcommand{\0}[1][]{\mathbf{0}_{#1}}
\newcommand{\I}[1][]{\mathbf{I}_{#1}}

\newcommand{\gyro}[1][]{{}^{#1}\tilde{\boldsymbol{{\omega}}}_t - {}^{#1}\mathbf{w}_{t}^{\omega}}

\newcommand{\gyroW}[1][]{{}^{#1}\mathbf{w}_{t}^{\omega}}

\newcommand{\acc}[1][]{{}^{#1}\tilde{{{\mathbf{a}}}}_t - {}^{#1}\mathbf{w}_{t}^{a}}

\newcommand{\Jth}[1][]{\mathbf{J}_{\theta}(\tilde{\mathbf{q}}_t)}